\documentclass[10pt,journal,compsoc]{IEEEtran}
\ifCLASSOPTIONcompsoc
  \usepackage[nocompress]{cite}
\else
  \usepackage{cite}
\fi
\usepackage[pdftex]{graphicx}
\ifCLASSINFOpdf
\else
\fi
\usepackage{amsmath}
\usepackage{ragged2e}
\usepackage{xcolor}
\usepackage{caption3}
\usepackage[normalem]{ulem}
 \usepackage[backref]{hyperref}

\begin{document}
%
\title{Sampling Equivariant Self-attention Networks \\for Object Detection in Aerial Images}
%
%
%
%

\author{Guo-Ye Yang,
        Xiang-Li Li,
        Ralph R. Martin,
        Shi-Min Hu,~\IEEEmembership{Senior Member,~IEEE}

\IEEEcompsocitemizethanks{\IEEEcompsocthanksitem 
G.-Y. Yang and X.-L. Li are with Tsinghua University,Beijing 100084, China.
\IEEEcompsocthanksitem 
R.-R. Martin was with the School of Computer Science and Informatics, Cardiff University, UK.
\IEEEcompsocthanksitem 
S.-M. Hu is with Tsinghua University, Beijing 100084, China. \\
E-mail: shimin@tsinghua.edu.cn}
\thanks{Manuscript received June 17, 2021.}}

%
%

\markboth{Journal of \LaTeX\ Class Files,~Vol.~14, No.~8, August~2015}%
{Shell \MakeLowercase{\textit{et al.}}: Bare Demo of IEEEtran.cls for Computer Society Journals}
%



\IEEEtitleabstractindextext{%

\justifying\let\raggedright\justifying
\begin{abstract}

Objects in aerial images have greater variations in scale and orientation than in typical images, so detection is more difficult.
Convolutional neural networks use a variety of frequency- and orientation-specific kernels to identify objects subject to different transformations; these require many parameters. Sampling equivariant networks can adjust  sampling from input feature maps according to the transformation of the object, allowing a kernel to extract features of an object under different transformations. Doing so requires fewer  parameters, and makes the network more suitable for representing deformable objects, like those in aerial images.
However, methods like deformable convolutional networks can only provide sampling equivariance under certain circumstances, because of the locations used for sampling.
We propose \emph{sampling equivariant self-attention networks} which consider self-attention restricted to a local image patch as convolution sampling with masks instead of locations, and design a transformation embedding module to further improve the equivariant sampling ability.
We also use a novel randomized normalization module to tackle overfitting due to limited aerial image data. 
We show that our model (i) provides significantly better sampling equivariance than existing methods, without additional supervision, (ii) provides improved classification on ImageNet, and (iii) achieves state-of-the-art results on the DOTA dataset, without increased computation.

\end{abstract}

}

\maketitle

\IEEEdisplaynontitleabstractindextext

%
\IEEEpeerreviewmaketitle

\IEEEraisesectionheading{\section{Introduction}\label{sec:introduction}}

\IEEEPARstart{O}{bject} detection in aerial images is a special case of the object detection task, playing an important role in fields such as environmental sciences, geosciences and ecology.
Compared to natural images, objects in aerial images have greater rotational and scale variations because of the overhead image capture, which requires the detection model to more flexibly handle geometric transformations. 

Convolutional neural networks (CNNs) have provided significant improvements for the object detection task. They use a variety of frequency- and orientation-specific kernels to identify objects under different transformations \cite{krizhevsky2012imagenet}, which requires many  parameters. They also must be trained with images of similar objects under various shooting conditions, leading to high training data acquisition costs.

To alleviate the above problems, networks with sampling equivariance adjust sampling of the input feature map in accordance with the object transformation, and extract features of objects with different transformations using the same set of convolutional parameters. Thus,  convolutional networks with sampling equivariance have greater feature extraction capabilities for the same number of parameters, and achieve better accuracy.
\begin{figure}[!t]        
 \center{\includegraphics[width=0.48\textwidth]  {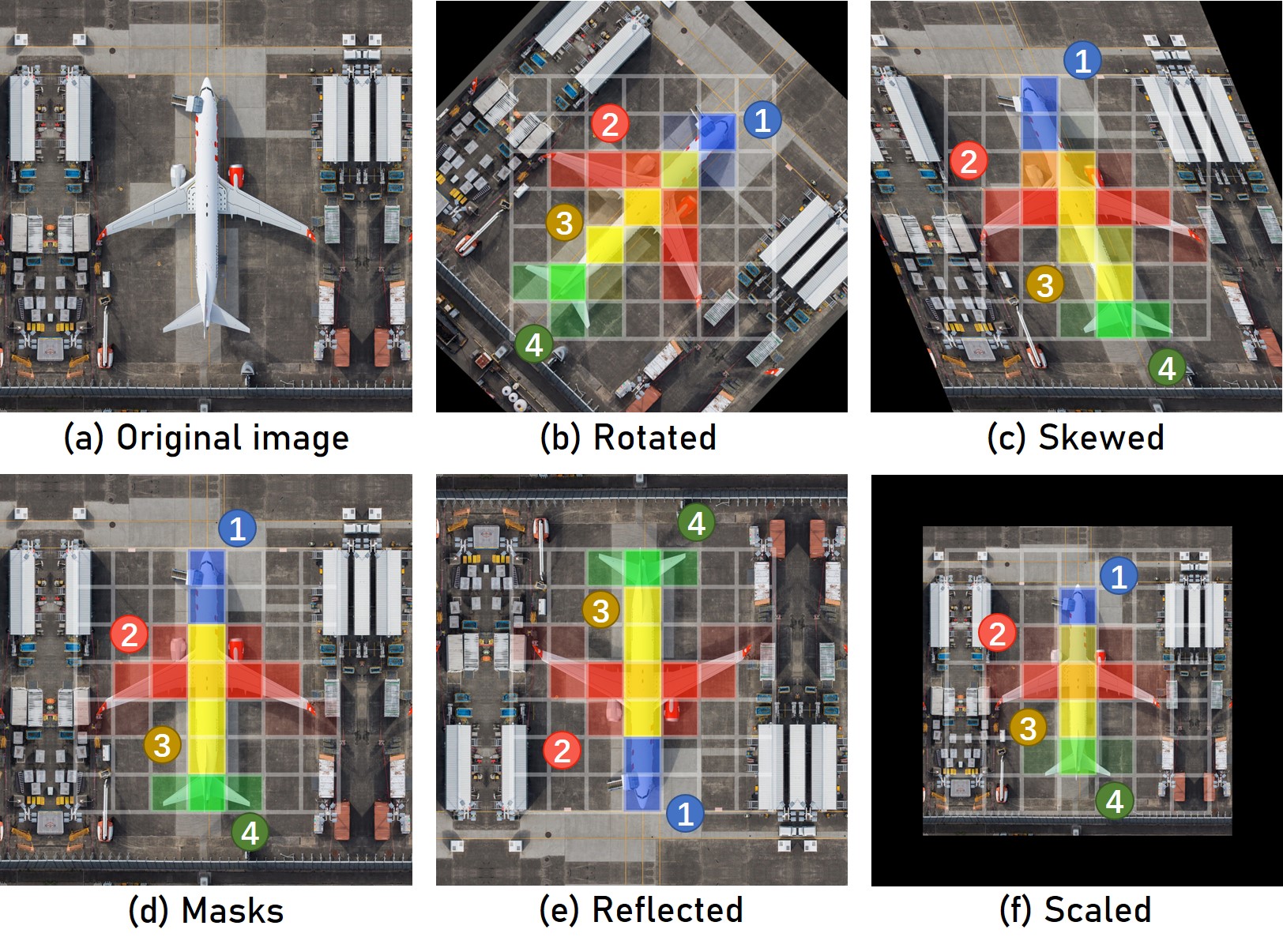}}        
 \caption{Illustration of equivariant sampling using masks. 
 We use four $7\times 7$ sampling masks here, colored blue, red, yellow and green. (a): original image. (d): sampling masks over the original image. (b, c, e, f): equivariant sampling masks for different transformations of the original image.
 }      
 \label{masksample}
\end{figure}

\begin{figure}[!t] 	
    \centering	
    \includegraphics[width=\linewidth]{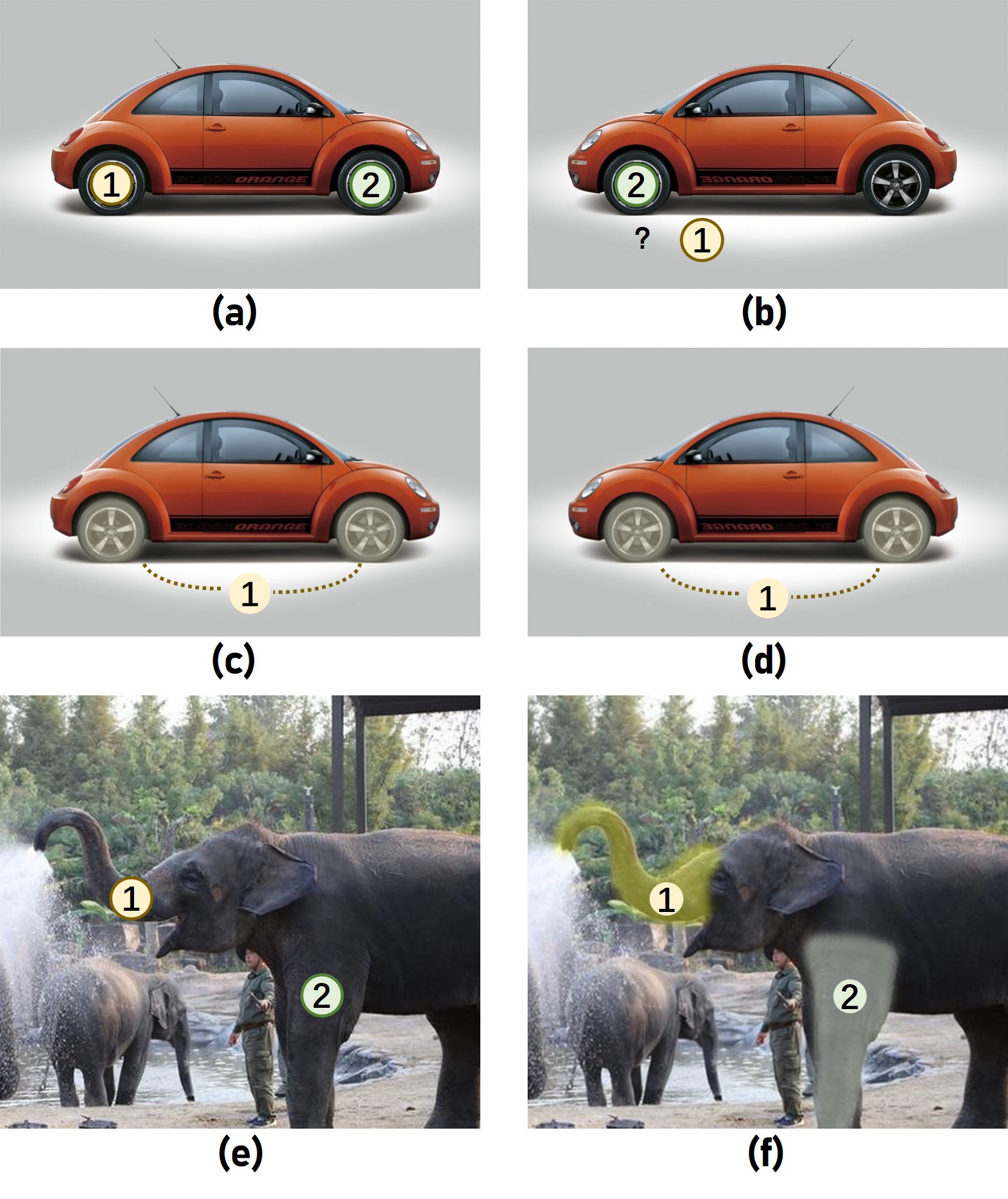} 	
    \caption{Sampling by location versus sampling by mask.
     Given similar-looking areas in the image, equivariant sampling by location may cause ambiguity under a reflection transformation  (a,b), while equivariant sampling by mask does not cause the problem (c,d). (e,f) show the advantage of masks over locations in expressing irregular areas.
	} 	
    \label{sampling} 
\end{figure}

Dai et al.~\cite{dai2017deformable} proposed deformable convolutional networks (DCNs) which obtain sampling equivariance by 
adding a regressive offset to each regular grid sampling location of a convolution.
However, DCN can only achieve sampling equivariance for scaling transformations, by correlating the receptive field of the deformable filters with object size, but does not provide  sampling equivariance under rotation, reflection, and skew. This results from the two drawbacks of sampling by location:
\begin{enumerate}
\item
 As Figs. \ref{sampling}(a) and \ref{sampling}(b) show, sampling  by location may cause confusion when  multiple image areas have similar appearances, and ambiguity may occur under transformations such as reflection. 
\item
As Fig. \ref{sampling}(e) shows, sampling represented by locations cannot  represent irregular areas well. 
\end{enumerate}

Self-attention based models are widely used in natural language processing and have recently been  adopted for vision tasks. They can perform position-independent feature aggregation based on the relationships between features, which is required for sampling equivariance.
To achieve general sampling equivariance, we  propose a novel sampling equivariant self-attention layer (SES-Layer). It considers self-attention restricted to a local image patch and extracts features using a sliding window as a convolution sampled by masks instead of locations.  As shown in Fig. \ref{sampling}(c--f), using masks for sampling can overcome the ambiguity of similar-looking regions, and can represent irregular areas better. Equivariant sampling by masks is illustrated in Fig.~\ref{masksample}. 

SES-Layer is implemented on top of self-attention networks (SANs)\cite{zhao2020exploring}, which do not have the property of sampling equivariance.
We improve SANs to meet the two requirements of sampling equivariance. First, we use a new positional encoding method to avoid the generation of  sampling masks being affected by sampling position, which would lead to a prior bias. 
Second, finding features after sampling may lose information of where and how the sampling is performed, e.g.\ how many wheels there are, where the wheels are, and what the shape of an irregular area is, etc. We propose to embed the sampling mask into the output features using a transformation embedding module to provide features with information related to the transformation of the sampling. Quantitative experiments show that our model achieves much better sampling equivariance without additional supervision, thus outperforming baseline models on the ImageNet dataset, without increasing the number of parameters or computation required.

Furthermore, acquisition of aerial images is costly and difficult, resulting in limited availability of training data; models based on self-attention are prone to overfitting on a small training set. To solve this problem, we propose a randomized normalization module (RNM) to enhance the generalization ability of the SES-Layer. Experiments show that this can significantly benefit network training when using a relatively small amount of data.
Combining this with sampling equivariance, our model outperforms state-of-the-art methods on the DOTA dataset (dataset for object detection in aerial images) without increased computation.

In summary, our contributions are as follows:
\begin{itemize}
\item  a sampling equivariance layer (SES-Layer), which  significantly improves sampling equivariance in the network,
\item a randomized normalization module (RNM) to enhance the generalization ability of the network when using a small training dataset, and
\item experimental verification that our proposed method outperforms baseline models on the ImageNet dataset and achieves state-of-the-art results on the DOTA dataset without additional computation.
\end{itemize}

\section{Related work}
\subsection{Transformation-Invariant Features}

Extracting transformation-invariant features from 2D images is an important problem in  computer vision.
Earlier work mostly used manually designed feature extractors~\cite{tuytelaars2000wide, mikolajczyk2001indexing, mikolajczyk2002affine, lowe2004distinctive, mikolajczyk2004scale, rublee2011orb}.
Lowe et al.~\cite{lowe1999object} proposed the scale-invariant feature transform (SIFT). 
Svetlana et al.~\cite{lazebnik2006beyond} proposed dividing an image into  patches, and used histograms of local features in all patches as the features of the image. 

In recent years, researchers began to use CNNs to extract transformation-invariant features. One class of methods achieves invariance for  specific transformations like translation, rotation, scaling and reflection by adding manually designed layers to the neural network~\cite{sohn2012learning, sifre2013rotation, kanazawa2014locally, laptev2016ti, worrall2017harmonic, bruna2013invariant, cohen2016group}.
Other methods use learnable deformation modules in the network, and
extract transformation-invariant features through indirect supervision of visual tasks such as classification and detection~\cite{yi2016lift,uhrig2017sparsity}.
Jaderberg et al.~\cite{jaderberg2015spatial} proposed spatial transformer networks (STN) which can perform a global affine transformation on the feature map according to the input, to get  invariant features.
Dai et al.~\cite{dai2017deformable} proposed deformable convolutional networks (DCNs) to provide the ability to transform local features, by adding an offset to each regular grid sampling location of the convolution. The offsets are regressed according to the input feature map.  
Some recent works improve DCNs by regressing the weights of the sampling locations according to  importance~\cite{zhu2019deformable} and sampling on the convolutional weights instead of feature maps~\cite{gao2019deformable}. 

Such methods~\cite{dai2017deformable, zhu2019deformable, gao2019deformable} obtain transformation invariance of local features by changing the sampling locations or weights of the convolution, which requires the sampling locations to have transformation equivariance. However, experiments show that they only have sampling equivariance for scaling transformations, and not for rotation, reflection, and skewing. 
We believe this is due to sampling by locations.
We propose to use masks for sampling instead of locations, which can overcome ambiguities, and better represent irregular areas to get better sampling equivariance.

\subsection{Self-Attention in Computer Vision}
    
With the wide success of the attention mechanism in natural language processing, similar techniques have been applied to computer vision, such as residual attention networks~\cite{wang2017residual}, squeeze-and-excitation networks~\cite{hu2018squeeze} and non-local neural networks~\cite{wang2018non}.
Transformer~\cite{vaswani2017attention} is ever more popular in natural language processing, and it has also been applied to computer vision.  Transformer has been demonstrated to have extraordinary performance due to its powerful representational capabilities, even surpassing CNNs on some tasks~\cite{dosovitskiy2020image,carion2020end,zhu2020deformable,Guo21pct}. Many transformer-based methods have been proposed for object detection~\cite{zheng2020end,dai2020up,sun2020rethinking}, image segmentation~\cite{wang2020max,wang2020end,ham,zheng2020rethinking}, image processing~\cite{zeng2020learning,zhou2018end} and pose estimation~\cite{huang2020hand,lin2020end,huang2020hot}.
Other works have been proposed to improve the efficiency of transformers, such as ASH~\cite{michel2019sixteen}, TinyBert~\cite{jiao2019tinybert}, FullyQT~\cite{prato2019fully}, ConvBert~\cite{jiang2020convbert} and ExternalAttention~\cite{guo2021attention}.

Some recent methods  apply self-attention to a local patch of the image~\cite{ramachandran2019stand,hu2019local,zhao2020exploring}.
Hu et al.~\cite{hu2019local} proposed a local relation layer for image recognition, which can adjust aggregation weights according to the relationships between local pixels. Zhao et al.~\cite{zhao2020exploring} explored variations of self-attention and proposed self-attention networks (SAN) with two forms of self-attention, pairwise and patchwise. 
In this paper, we regard self-attention restricted to a local image patch as a convolution sampled by mask instead of location.
SAN-patchwise uses different weights to regress the masks in different sampling locations, which leads to inequality of different sampling locations, making it hard to obtain sampling equivariance. Thus, our proposed model is implemented on top of SAN-pairwise, which we also use as one of our baseline models.

Our proposed model has significantly better sampling equivariance compared to the baselines, and we further provide a randomized normalization module (RNM) to solve the problem of overfitting of self-attention models trained on small datasets.

\subsection{Object detection in aerial images}
Although object detection has been investigated for decades, there are still  unique challenges for aerial images. Object detection in aerial images faces problems of varying scale and orientation, as well as densely packed objects. 
General purpose object detection models cope poorly with these aspects of objects in aerial images, and datasets based on unspecialised images are poorly suited for use in training aerial image detection models. Thus many aerial datasets and aerial object detection methods have been proposed.

Typical aerial image object detection datasets are annotated with different objects, such as ships~\cite{liu2016ship}~\cite{chen2020fgsd}, planes~\cite{zhu2015orientation}, buildings~\cite{benedek2011building}, vehicles~\cite{razakarivony2016vehicle,liu2015fast,mundhenk2016large}, or multiple categories~\cite{xia2018dota,ding2021object}.  DOTA~\cite{xia2018dota} is one of the most popular aerial image datasets, with 2806 aerial images from different sensors and platforms; it contains 188,282 object instances labeled in 15 common categories. We use DOTA for evaluation and comparison in our experiments.
 
To detect objects of uncertain orientation in aerial images, many works have suggested how to learn in a rotation-invariant way~\cite{cheng2016learning,cheng2016rifd, cheng2018learning,liu2017learning}, while~\cite{liu2017rotated,zhang2018toward,yang2018automatic} proposed many oriented detectors for ship detection in aerial images. In order to make better use of  deep features for detecting objects with uncertain orientation, Zhou et al.~\cite{zhou2017oriented} proposed oriented response networks to produce within-class rotation-invariant deep features, and Wang et al.~\cite{wang2020learning} hoped to learn center probability to improve  performance. By considering multiple scales of objects in aerial images, Azimi et al.~\cite{azimi2018towards} improved detection of such objects by using FPN and deformable convolutions.
Ding et al.~\cite{ding2019learning} proposed a region of interest transformer that can turn an axis aligned bounding box into an oriented bounding box. Li et al.~\cite{li2019learning} proposed an object-wise semantic representation to improve object detection, while Zhang et al.~\cite{zhang2019cad} exploited scene-level global contextual features and an attention module to improve detection. SCRDet~\cite{yang2019scrdet} and SCRDet++~\cite{yang2020scrdet++}  improve  detection by use of an attention mechanism, instance-level feature denoising and rotation loss smoothing.
For faster and more accurate oriented object detection, Yang et al.~\cite{yang2019r3det} proposed a refined single-stage rotation detector. Since focal loss~\cite{lin2017focal} was proposed for dense object detection, many different oriented bounding box representations and regression losses have been proposed, such as circular smooth labels~\cite{yang2020arbitrary}, modulated loss~\cite{qian2019learning}, dense label encoding~\cite{yang2020dense}, and Gaussian Wasserstein distance loss~\cite{yang2021rethinking}. Pan et al.~\cite{pan2020dynamic} proposed a feature selection module and a dynamic refinement head to adjust the receptive field adaptively and dynamically refine  object prediction. Han et al.~\cite{han2021align} proposed a single-shot alignment network and achieved state-of-the-art performance by aligning deep features adaptively and reducing inconsistency between localization and classification. 
 
Xu et al.~\cite{xu2020gliding} proposed a novel representation for an oriented bounding box by four vertices `gliding' on the four sides of an axis-aligned bounding box (AABB); they regressed the AABB and the length ratios which represent the offset along each corresponding side of the oriented bounding box. 
The resulting gliding model achieves state-of-the-art performance on DOTA. Because this network is concise and achieves good results, it is a suitable baseline for measuring the effect of our proposed SES-Layer in aerial image object detection.
\section{Methodology}
\begin{figure*}[!t]
	\centering
	\includegraphics[width=1\textwidth]{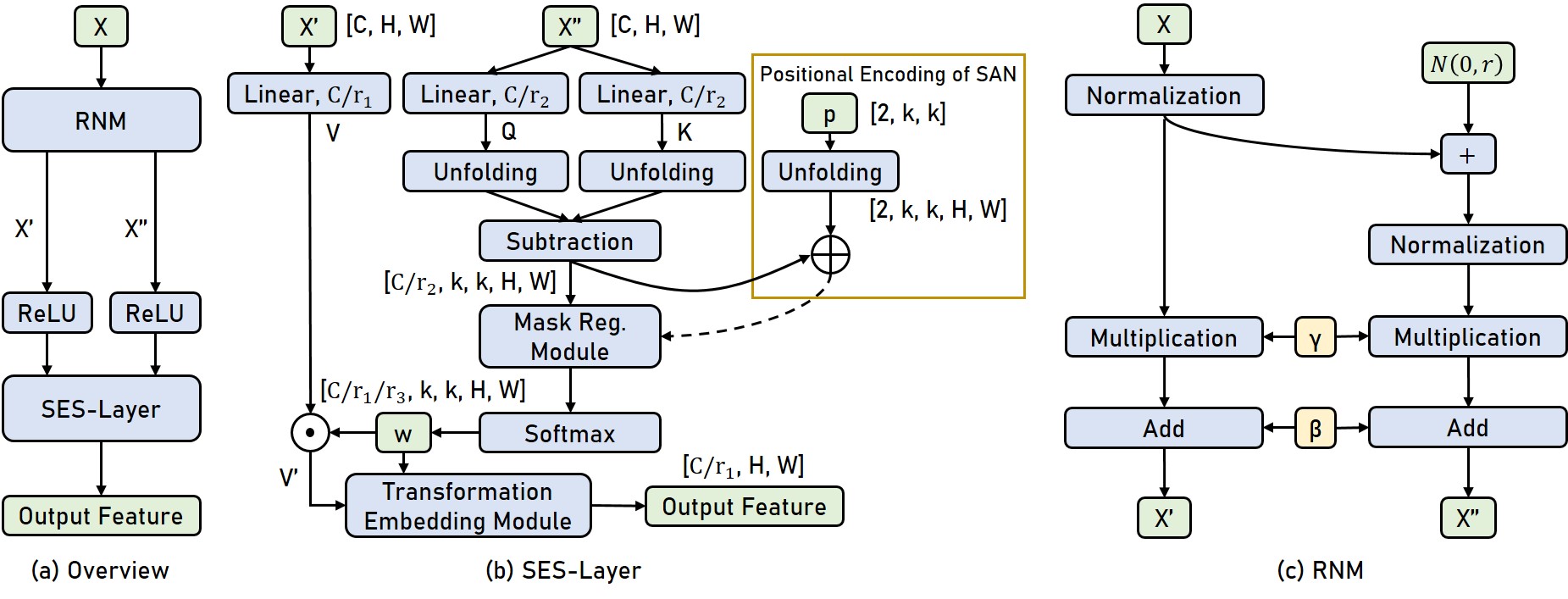}
	\caption{Proposed network. (a): Overall network comprising RNM, ReLUs, and SES-Layer. (b) Sampling equivariant self-attention layer (SES-Layer); $\bigodot$ denotes Hadamard product, and $\bigoplus$ denotes concatenation. Yellow box, right: removed positional encoding of SAN. (c) Randomized normalization module (RNM).}
	\label{pipline}
\end{figure*}

\subsection{Architecture}
Our proposed network is shown in Fig. \ref{pipline}. It has two main  components, a randomized normalization module (RNM) and a sampling equivariant self-attention layer (SES-Layer). The SES-layer considers self-attention restricted to a local patch of the image and extracts features with a sliding window as a convolution sampled by masks, instead of sampling by location as in previous work~\cite{dai2017deformable,zhu2019deformable}. The RNM is designed to enhance the generalization ability of the SES-Layer when trained on  small  datasets.


\subsection{Sampling Equivariant Self-attention Layer}
Our analysis of  why previous methods~\cite{dai2017deformable, zhu2019deformable, hu2019local,ramachandran2019stand, zhao2020exploring} failed to achieve sampling equivariance led us to propose a sampling equivariant self-attention layer (SES-Layer) based on pairwise self-attention networks (SAN)~\cite{zhao2020exploring}; the pipeline is shown in Fig. \ref{pipline}(b).

$X''$ and $X'$ are  feature maps with and without random perturbations. They have shape $[C, H, W]$, where the $C$, $H$ and $W$ represent the number of channels, height and width of the feature map. $X'$ and $X''$ are fed into three linear layers to get the three output tensors \emph{value}, \emph{query} and \emph{key},  denoted  $V$, $Q$ and $K$, where $r_1$, $r_2$ are bottleneck dimension reduction factors, and $r_3$ is the mask sharing factor.
The sampling mask $w$ is obtained using:
\begin{equation}
w=\mathrm{softmax}(\gamma(\delta(Q)-\delta(K)))
\end{equation}
where $\delta$ is an unfolding operator with kernel size $k$, and $\gamma$ is a mask regression module composed of [BN, ReLU, Linear, BN, ReLU, Linear] (BN being batch normalisation). The linear layer operates on the first dimension of the feature, and the softmax layer operates on the second and third dimensions of the feature. 
In SAN, features need to be position encoded before entering the mask regression module $\gamma$, but we remove this step.

After obtaining the sampling mask $w$, the output feature map $Y$ is obtained as:
\begin{equation}
Y=\zeta(V\odot w, w)
\end{equation}
where $\odot$ denotes the Hadamard product, and $\zeta$ is a transformation embedding module with two inputs: the features after equivariant sampling $V'=V\odot w$ and the sampling masks $w$. 
In the transformation embedding module, we take the feature for each sample in $V'$ and the sampling mask corresponding to this feature with a shape of $k\times k$ as the sampling transformation information. Then, a linear layer with shared weights is used to embed the transformation information into the feature. Shared weights are used as we believe different features can embed transformation information in the same way, and doing so makes  training  the model easier. In addition, the use of shared weights  reduces the number of parameters, resulting in reduced computation. Compared to positional encoding, the transformation embedding module does not result in a greater number of parameters or computational workload.

Positional encoding can be used to provide spatial context for the model, but it causes prior bias in the generated sampling masks. We thus remove the positional encoding and embed the sampling mask $w$ into the features after equivariant sampling. This operation retains the function of providing spatial context for positional encoding while removing the prior error, and ensures that the features after equivariant sampling have the sampling transformation information, which simultaneously overcomes both reasons that SAN cannot achieve sampling equivariance. In our experiments, we find that the sampling equivariance of our proposed model is significantly improved, as has the performance of the model on different tasks.

The SES-Layer can simply replace any convolution layer with kernel size greater than 1 in a CNN, to give it  sampling equivariance and allowing more efficient extraction of features. 

\subsection{Randomized Normalization Module}
A network based on self-attention requires a large amount of training data to generalize well~\cite{dosovitskiy2020image}. However, due to the cost and difficulty of aerial image acquisition, the amount of available training data is relatively small~\cite{xia2018dota}. We thus instead hope to enhance the generalization ability of our proposed model by adding random noise to the input of the SES-Layer. However, adding noise directly to the input feature map will cause the data distribution to be inconsistent between training and testing, reducing performance. 
To solve this problem, we propose a randomized normalization module (RNM) based on batch normalization (BN), and explore how to combine it with the SES-Layer. The pipeline of the RNM is shown in Fig. \ref{pipline}(c).

During training, for an input feature map $X$ of shape $[C, H, W]$, the RNM  produces two outputs, the first being the output $X'$ of typical BN:
\begin{equation}
X'=\gamma(X-\mu_{X})/(\sigma_{X}^{2}+\epsilon)+\beta
\end{equation}
where $\mu_{X}$ denotes the mean of $X$ across channels, $\sigma_{X}^{2}$ denotes the variance of $X$ across channels, $\gamma$ and $\beta$ are learnable weights for BN with shape $[C]$, and $\epsilon$ is set to 0.00001.
The second output of RNM is a randomized output $X''$ computed using:
\begin{equation}
    \begin{aligned}
        \hat{X}&=X+N(0,r)\\
        X''&=\gamma(\hat{X}-\mu_{\hat{X}})/(\sigma_{ \hat{X} }^{2}+\epsilon)+\beta
    \end{aligned}
\end{equation}
where $N(0,r)$ represents a normal distribution generator with output shape $[C, H, W]$, mean $0$ and variance $r$.

After applying ReLU to the feature maps $X'$ without random perturbation  and $X''$ with random perturbation, we take $X'$ as the input $V$, and $X''$ as inputs $Q$ and $K$, in the SES-Layer: we believe that overfitting the \emph{query} and \emph{key} in the self-attention structure makes it difficult to generate more generalized sampling masks during inferencing, so a random disturbance should be added to the \emph{query} and \emph{key} inputs. However, random disturbance of the \emph{value} input  will increase noise in the features, which is not conducive to  training. Our experiments demonstrate that this randomization approach is the one most conducive to improving the generalization ability of the model. RNM behaves like BN during inferencing, so does not result in additional computation, nor additional parameters.

\subsection{Sampling Equvariance}
\subsubsection{Analysis of Sampling Methods}

We now analyze convolution, deformable convolutional networks (DCNs)~\cite{dai2017deformable}, local self-attention models, and our proposed SES-Net in terms of  sampling approach, and discuss their effects on sampling equivariance and model performance.

\textbf{Convolution} uses fixed regular grid locations to sample the input feature map. This structure is relatively inflexible and requires more parameters to accommodate different deformations.

\textbf{DCNs}~\cite{dai2017deformable,gao2019deformable,zhu2019deformable} use locations determined by regression for sampling, which gives the model a certain degree of flexibility. Experimentally, we find that this structure can achieve good sampling equivariance under a scaling transformation, correlating the receptive field of the DCNs  with object size; this property is also visualized in~\cite{dai2017deformable}. Thus, a DCN convolution kernel   can extract features of objects of different sizes, reducing the number of parameters needed. However,
as noted in Sec. 1 and shown in Fig. \ref{sampling}, because the use of locations for sampling causes confusion, it cannot achieve complete sampling equivariance.

\textbf{Local self-attention models}\cite{ramachandran2019stand,hu2019local,zhao2020exploring} apply self-attention to local patches of the image, and use sliding windows to extract features. These works do not consider local self-attention in terms of sampling with masks, and thus their models have drawbacks that make it difficult to achieve sampling equivariance.

\textbf{SES-Net} regards aggregation of mask $w$ and $V$ in the self-attention structure as a sampling process in $V$ with mask $w$. From this perspective,  existing methods can be considered to have problems of sampling prior error and lack of sampling transformation information. SES-Net overcomes them with a transformation embedding module. 
Sampling using masks has greater flexibility than using locations and does not cause confusion. After solving these two problems, significantly better sampling equivariance can be achieved under more diverse transformations.
A model with sampling equivariance can use a single kernel to extract features from objects subject to different transformations, which can give the model much stronger feature extraction capabilities. 
Our proposed model can thus achieve better accuracy with the same number of parameters.

\subsubsection{Sampling Equivariance Strategy Evaluation}
In order to objectively compare the sampling equivariance of models with different sampling methods,
we have designed a unified quantitative evaluation metric. 

SES-Net and SAN use masks to sample  feature maps, while DCNs use locations to sample through bilinear interpolation. Since the feature of each sampling can be calculated as the weighted average of each cell in a feature map,  we can construct a sampling graph of sampling locations and sampling masks using the weights. The earth mover's distance (EMD)~\cite{rubner2000earth} is a measure of the distance between two probability distributions, and is a good metric to measure the distance between two sampling graphs. Thus, we use EMD to quantitatively evaluate the equivariance of sampling locations and masks before and after transformation.

To evaluate sampling equivariance on transformation $T$, we take $N$ images. For the i-th image $I_i$, we first randomly select transformation parameters for $T$ to obtain the transformation $T_i$, and compute the transformed image $\widehat{I}_i$ by applying $T_i$ to $I_i$. During feature extraction, there are $M_{d}$ feature maps with stride $d$. We first randomly select a stride $d_i$, then feed $I_i$ and $\widehat{I}_i $ into the network to get the $j$-th feature map for stride $d_i$, and take the $k$-th sampling graph in this feature map $g_{i,j,k}$ and $\widehat{g}_{i,j,k}$. Finally, we use $T_i$ to transform $g_{i,j,k}$ to obtain an ideally equivariant sampling graph $\widetilde{g}_{ i,j,k}$, and use average EMD (AEMD) to evaluate the sampling equivariance:
\begin{equation}
\mathrm{AEMD}=\alpha\frac{1}{N}\sum_{i=1}^{N}\frac{1}{M_{d_i}}\sum_{j=1}^{M_{d_i}}\frac{1}{O_{d_{i},j}}\sum_{k=1}^{O_{d_{i},j}}\mathrm{EMD}(\widetilde{g}_{i,j,k}, \widehat{g}_{i,j,k})
\label{eAEMD}
\end{equation}
where $\alpha=1/112$ is a scalar to normalize the theoretical maximum value of AEMD to 1. A smaller value indicates better sampling equivariance, and $O_{d_{i},j}$ indicates the number of samplings at a sampling center on the $j$-th feature map of stride $d_i$.

\section{Experiments}
\subsection{Approach}
We first conducted a series of experiments on the ImageNet~\cite{deng2009imagenet} dataset to evaluate the sampling equivariance and  performance of SES-Net and baseline models. 
Then we conducted experiments on DOTA~\cite{xia2018dota} to show the importance of sampling equivariance in object detection in aerial images. The results demonstrate that, by taking advantage of  sampling equivariance, our network can provide improved accuracy over the state-of-the-art model while reducing the number of parameters and computation needed. Finally, we verified the effectiveness of our proposed modules through ablation studies.
Our method is built on top of SAN-pairwise, which we thus regard as one of the baseline models; we denote it SAN in the following experiments. 

\subsection{Datasets and Evaluation Metrics}
\subsubsection{ImageNet~\cite{deng2009imagenet}} ImageNet is a large-scale hierarchical image dataset spanning 1,000 object classes and containing 1,281,167 training images, 50,000 validation images and 100,000 testing images. We use the training set for training and the validation set for testing. In the testing stage, we use top-1 and top-5 classification accuracy as the evaluation metric.

\subsubsection{DOTA~\cite{xia2018dota}} DOTA is a popular aerial image dataset for oriented object detection. It has 2806 aerial images of sizes from   $800 \times 800$ to $4000 \times 4000$ from different sensors and platforms. It contains 188,282 instances labeled in 15 common categories: Plane (PL), Ship (SH), Storage Tank (ST), Baseball Diamond (BD), Tennis Court (TC), Basketball Court (BC), Ground Track Field (GTF), Harbor (HA), Bridge (BR), Large Vehicle (LV), Small Vehicle (SV), Helicopter (HC), Roundabout (RA), Soccer-Ball Field (SBF) and Swimming Pool (SP). Following the standard protocol, we use the training set and validation set for training and the testing set for testing, and use mean average precision (mAP) as the evaluation metric.

\begin{table}[!t]
	\centering
	\caption{ Average earth mover's distance (AEMD) in sampling equivariance experiments. }
	\label{AEMD}
	\renewcommand\tabcolsep{5pt}
	\begin{tabular}{lrrrr}
		\hline\noalign{\smallskip}
		Method & Rotation $\downarrow$ & Reflection $\downarrow$ & Skew $\downarrow$ & Scale $\downarrow$\\
		\noalign{\smallskip}
		\hline
		\noalign{\smallskip}
        SAN10 & 0.0122 & 0.0068 & 0.0067 & 0.0072
        \\
		SES-Net10 (Ours) & 0.0065 & 0.0032 & 0.0063 & 0.0058
        \\
		\noalign{\smallskip}
		\hline
		\noalign{\smallskip}
        SAN15 & 0.0110 & 0.0057 & 0.0067 & 0.0071
        \\
        SES-Net15 (Ours) & 0.0061 & 0.0029 & 0.0061 & 0.0057
		\\
		\noalign{\smallskip}
		\hline
		\noalign{\smallskip}
        DCNv2 & 0.0293 & 0.0243 & 0.0225 & 0.0068
        \\
        DCNv2-C & 0.0267 & 0.0215 & 0.0199 & 0.0063
        \\
        SAN19 & 0.0099 & 0.0050 & 0.0062 & 0.0065
        \\
        SES-Net19 (Ours) & \textbf{0.0059} & \textbf{0.0028} & \textbf{0.0059} & \textbf{0.0055}
        \\
		\hline
	\end{tabular}
\end{table}

\begin{figure*}[!t]
	\centering
	\includegraphics[width=\textwidth]{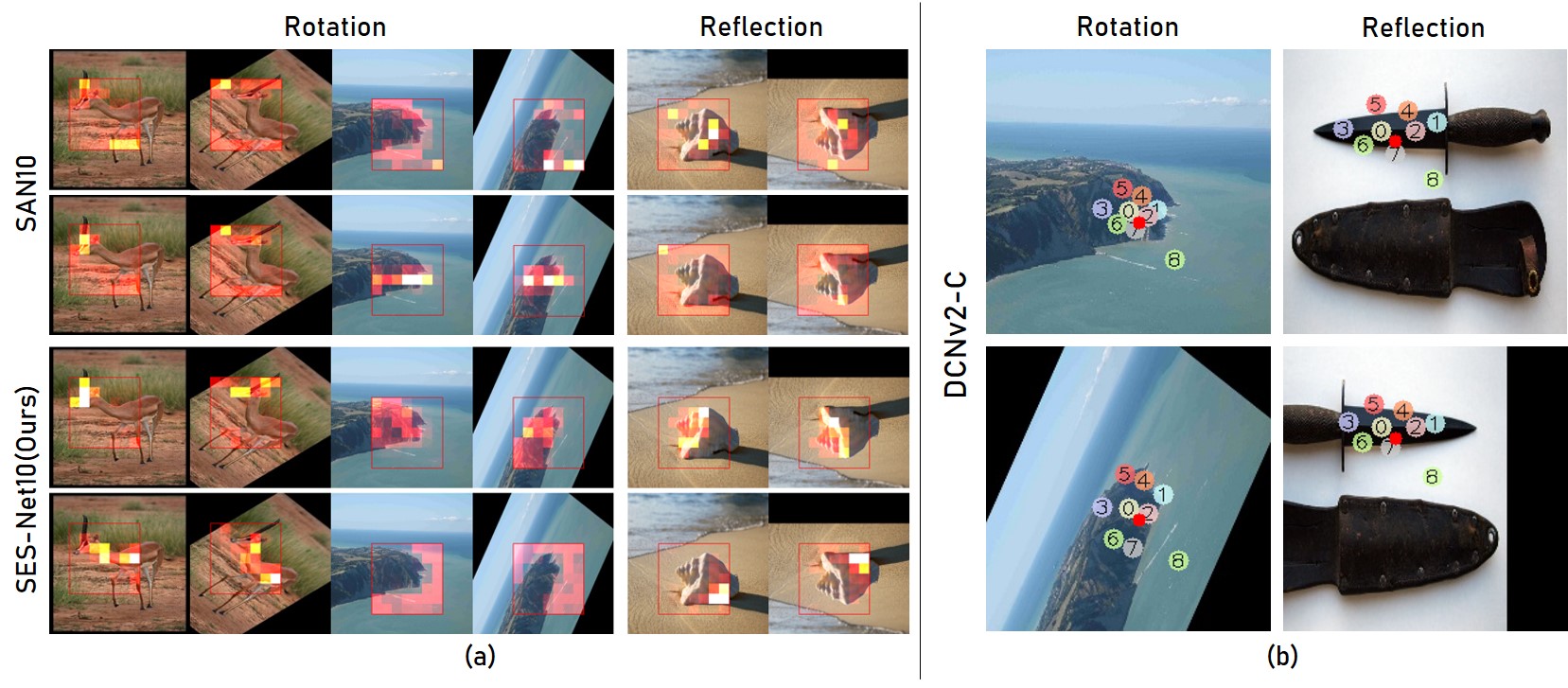}
	\caption{
	Samples for different methods for images before and after transformation.
	Columns: before and after different transformations: (a1--a4, b1): rotation, (a5, a6, b2): reflection. Rows: different methods:  (a1, a2): SAN10, (a3, a4): SES-Net10, where for each image, two example samplings are shown  with each sampling mask in the original image, (b1--b2): DCNv2-C, where  numbered circles represent different sampling locations.}
	\label{attention}
\end{figure*}

\subsection{Backbone Performance}
We next describe experiments on sampling equivariance and the ImageNet classification task to validate the effectiveness of our proposed model as a backbone network structure.

\subsubsection{Implementation details}
Experiments in this section were conducted on the ImageNet~\cite{deng2009imagenet} dataset.
Our SES-Net10, SES-Net15 and SES-Net19 are  based on  SAN10, SAN15, and SAN19 respectively, replacing  SA layers with SES layers.
We conducted experiments on ImageNet without the RNM module, as there is sufficient training data.

In our experiments, the bottleneck dimension reduction factors and mask sharing factor were set to $r_1=4$, $r_2=16$ and $r_3=8$,  as for SAN. To ensure a fair comparison, we used the same data augmentation strategy as SAN,  including random cropping to $224\times 224$ patches and random horizontal flipping. Our models were trained for 100 epochs using a stochastic gradient descent (SGD) optimizer with batch size $256$, momentum $0.9$ and weight decay $0.0001$. The learning rate was initially set to $0.1$, with cosine annealing schedule. Our models were implemented using Jittor~\cite{hu2020jittor}.

We used SAN~\cite{zhao2020exploring}, ResNet~\cite{he2016deep} and DCNv2~\cite{zhu2019deformable} with ResNet-50 as backbone as our baseline models, and train SES-Net, SAN, ResNet and DCNv2 from scratch on ImageNet with the above hyper-parameters for classification, where DCNv2 used for classification is denoted  DCNv2-C. In addition, we also compared with DCNv2 with  weights provided by mmdetection~\cite{mmdetection} trained on MSCOCO~\cite{lin2014microsoft}.

\subsubsection{Sampling Equivariance}
In order to evaluate the sampling equivariance of different methods, we designed quantitative and qualitative experiments on sampling equivariance, and performed comparisons with SES-Net, SAN~\cite{zhao2020exploring} and DCNv2~\cite{zhu2019deformable}.

We considered four typical transformations: rotation, reflection, skewing and scaling. For each transformation, we randomly selected $N=5000$ images from the ImageNet validation set  for the experiment. We used AEMD from Equ. \ref{eAEMD} to evaluate the sampling equivariance of DCNv2, DCNv-C, SAN10, SAN15, SAN19, SES-Net10, SES-Net15 and SES-Net19; results are shown in  Tab. \ref{AEMD}. Compared to DCNv2, our results are significantly better
under rotation, reflection and skewing transformations, while DCNv2 has a sampling equivariance  closer to that of our model under scaling; it is hard for DCNv2 to achieve complete sampling equivariance by using locations for sampling. 
Our model also outperforms SAN for each task, as our proposed structure can eliminate SAN's prior sampling errors by retaining the spatial context and incorporating sampling transformation information into the output features.

In Fig. \ref{attention}, we show sampling examples on the ImageNet validation set using different methods. 
Due to space limitations, we only show results for DCNv2-C, SAN10 and SES-Net10 under rotation and reflection. 
Fig. \ref{attention}(b) shows results for DCNv2-C, numbered circles representing different sampling locations for DCNv2. 
Comparing locations 2, 4, and 7 in images in the first column shows that these locations have not rotated as the image rotates: these locations do not achieve sampling equivariance.
Rows (a1, a2) show sampling masks for SAN10. Due to prior errors and lack of transformation information, they do not always provide good sampling equivariance: see row a2, columns 3--4.
Rows (a3, a4) show sampling masks for our method, which provides good sampling equivariance under different image transformations. 
Both quantitative and qualitative experiments show our proposed model  provides significantly better sampling equivariance without special supervision.

\subsubsection{Classification}
\begin{table}[!t]
	\begin{center}
		\caption{Results using the  ImageNet test set.}
		\label{ImageNet}
		\begin{tabular}{lrrrrrrr}
			\hline\noalign{\smallskip}
			    Method & Top-1(\%) & Top-5(\%) & Params & FLOPs \\
			\noalign{\smallskip}
			\hline
			\noalign{\smallskip}
        		ResNet26 & 73.6 & 91.7 & 13.7M & 2.4G\\
        		SAN10 & 75.1 & 92.3 & 10.5M & 2.1G\\
        		SES-Net10 (Ours) & \textbf{76.4} & \textbf{93.0} & 10.6M & 2.1G\\
        		\noalign{\smallskip}
        		\hline
        		\noalign{\smallskip}
        		ResNet38 & 76.0 & 93.0 & 19.6M & 3.2G\\
        		SAN15 & 76.6 & 93.1 & 14.1M & 2.9G\\
        		SES-Net15 (Ours) & \textbf{77.5} & \textbf{93.7} & 14.2M & 3.0G\\
        		\noalign{\smallskip}
        		\hline
        		\noalign{\smallskip}
        		ResNet50 & 76.9 & 93.5 & 25.6M & 4.1G\\
        		DCNv2-C & 75.5 & 92.6 & 26.4M & 4.3G\\
        		SAN19 & 76.9 & 93.4 & 17.6M & 3.7G\\
        		SES-Net19 (Ours) & \textbf{77.8} & \textbf{93.8} & 17.8M & 3.8G\\
			\hline
		\end{tabular}
	\end{center}
\end{table}

In order to quantitatively evaluate the capability of our proposed model, we conducted experiments also using ResNet26, ResNet38, ResNet50, DCNv2-C, SAN10, SAN15, SAN19, SES-Net10, SES-Net15 and SES-Net19 for classification on ImageNet. As shown in Tab. \ref{ImageNet}, our method has better top-1 and top-5 accuracy than ResNet, using fewer parameters and FLOPs. Compared to SAN, top-1 and top-5 accuracy are also improved, using a similar number of parameters and FLOPs. It shows that the sampling equivariance of our proposed method enables the sampling to adjust to object transformations, allowing the model to extract features of objects under different transformations with the same parameters. Therefore, the model has stronger feature extraction capabilities, and achieves better performance.

\begin{table*}[!t]
	\begin{center}
		\caption{
		 Detection accuracy on different objects (AP) and overall performance (mAP) evaluation on the DOTA test set. }
		\label{dota}
		\renewcommand\tabcolsep{4.0pt}
		\scalebox{0.945}{
		\begin{tabular}{llrrrrrrrrrrrrrrrrrrrrrrr}
			\hline\noalign{\smallskip}
			    Method & Backbone & PL &BD & BR & GTF & SV & LV & SH & TC & BC & ST & SBF & RA & HA & SP & HC & mAP \\
			\noalign{\smallskip}
			\hline
			\noalign{\smallskip}
			    FR-O & R-101 & 79.42 & 77.13 & 17.70 &64.05& 35.30& 38.02 & 37.16 & 89.41 & 69.64 & 59.28 & 50.30 & 52.91 & 47.89 & 47.40& 46.30 & 54.13 \\
			    Azimi et al.  & R-101-FPN&81.36&74.30&47.70&70.32&64.89&67.82&69.98&90.76&79.06&78.20&53.64&62.90&67.02&64.17&50.23&68.16 \\
			    CADNet & R-101-FPN&87.80&82.40&49.40&73.50&71.10&63.50&76.60&90.90&79.20&73.30&48.40&60.90&62.00&67.00&62.20&69.90 \\
			    DRN & H-104&88.91&80.22&43.52&63.35&73.48&70.69&84.94&90.14&83.85&84.11&50.12&58.41&67.62&68.60&52.50&70.70
			    \\
			    R$^3$Det & R-101-FPN&89.54&81.99&48.46&62.52&70.48&74.29&77.54&90.80&81.39&83.54&61.97&59.82&65.44&67.46&60.05&71.69\\
			    SCRDet & R-101-FPN&89.98&80.65&52.09&68.36&68.36&60.32&72.41&90.85&\textbf{87.94}&86.86&65.02&66.68&66.25&68.24&65.21&72.61 \\
			    CenterMap &R-101-FPN&89.83&84.41&54.60&70.25&77.66&78.32&87.19&90.66&84.89&85.27&56.46&69.23&74.13&71.56&66.06&76.03 \\
			    Li et al. & R-101-FPN&\textbf{90.41}&85.21&55.00&78.27&76.19&72.19&82.14&90.70&87.22&86.87&66.62&68.43&75.43&72.70&57.99&76.36 \\
        		Gliding & R-101-FPN & 89.47 &84.63& 51.00 & 74.91& 71.56 & 74.28& 86.73& \textbf{90.86}& 79.55 & 86.78& 58.68& 70.65& 72.96& 70.76& 59.42 & 74.82 \\
        		Gliding+SES (Ours) & R-101-FPN& 89.41& \textbf{86.83}& 51.56& 73.22& 72.59& 75.46& 87.32& 90.59& 87.04& 86.83& 64.54& \textbf{70.68}& 74.91&74.21& 67.32 & 76.84\\
        		S$^2$ANet &R-50-FPN & 88.89 & 83.60 & \textbf{57.74} & 81.95 & 79.94 & 83.19 & 89.11 & 90.78 & 84.87 & 87.81 & \textbf{70.30} & 68.25 & 78.30 & \textbf{77.01} & 69.58 & 79.42 \\
        		S$^2$ANet+SES (Ours) & R-50-FPN& 88.96&82.99&56.99&\textbf{82.84}&\textbf{80.75}&\textbf{85.89}&\textbf{89.35}&90.58&86.67&\textbf{87.88}&67.96&69.23&\textbf{78.75}&73.04&\textbf{75.19} & \textbf{79.80}\\
			\hline
		\end{tabular}}
	\end{center}
\end{table*}

\begin{table}[!t]
	\begin{center}
		\caption{Number of backbone parameters and FLOPs for different methods. }
		\label{para}
		\begin{tabular}{lrr}
			\hline\noalign{\smallskip}
			    Method & B-Params & B-FLOPs \\
			\noalign{\smallskip}
			\hline
			\noalign{\smallskip}
        		Gliding & 44.5M & 7.8G\\
        		Gliding+SES(Ours) & 38.5M & 7.6G\\
        		S$^2$ANet & 25.6M & 4.1G\\
        		S$^2$ANet+SES(Ours) & 19.5M & 3.9G\\
			\hline
		\end{tabular}
	\end{center}
\end{table}

\begin{figure*}[!t]
	\centering
	\includegraphics[width=1\textwidth]{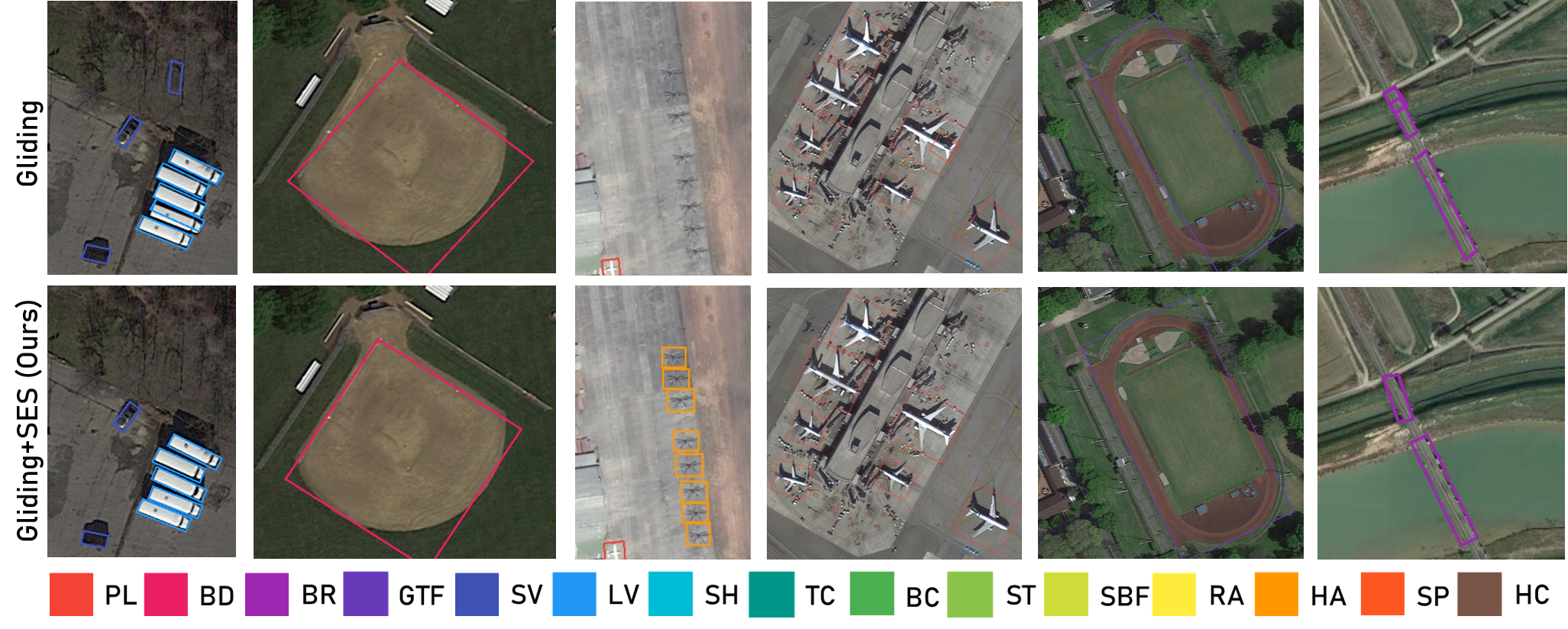}
	\caption{Detection results for different methods on DOTA images.
	Boxes in different colors show detected objects in different categories.
	Above: using gliding as the baseline method. Below: using our proposed Gliding+SES. }
	\label{dotaexp}
\end{figure*}

\subsection{Experiments on DOTA}
Object detection in aerial images differs from that in natural images because it needs to identify objects from a bird's-eye view. Furthermore, objects in aerial images can have large variations in scale and orientation, which must be handled by the model. Our proposed SES-Layer can give the model better sampling equivariance, enhancing its ability to deal with different transformations. To demonstrate this, we applied our proposed architecture to existing approaches, and conducted comparative experiments on the object oriented bounding box (OBB) detection task in DOTA.

\subsubsection{Implementation details}
We applied our SES-Layer and RNM to the gliding approach~\cite{xu2020gliding} with ResNet101 as backbone and to S$^2$ANet~\cite{han2021align} with ResNet50 as backbone, by simply replacing [BN, ReLU, Conv$3\times 3$] with [RNM, ReLU, SES-Layer] in the bottleneck of the backbone; these are denoted  Gliding+SES and S$^2$ANet+SES respectively. 
We set the random size to $r=0.005$ for Gliding+SES and $r=0.01$ for S$^2$ANet+SES, and set the kernel size $k=7$ for all SES-Layers. We set the dimension reduction factor $r_1=1$ in order to adapt to the structure of ResNet. 
Due to the limited amount of training data in DOTA, we only replaced bottlenecks in the last stage of ResNet to ease training. In the training process, we used parameters pre-trained on ImageNet as the initial weights. For Gliding+SES, the learning rate was initially set to 0.1 decayed by a cosine annealing schedule, and the other hyper-parameters were set as for Gliding. For S$^2$ANet+SES, we used the same hyper-parameters as S$^2$ANet. 

\subsubsection{Experiments }
We conducted experiments for the OBB task using DOTA. Since most previous works~\cite{cheng2016learning,zhou2017oriented,liu2017rotated,liu2017learning,zhang2018toward,yang2018automatic,ding2019learning,yang2019scrdet,yang2019r3det,han2021align} designed  special modules to handle rotation and scaling, it is hard to measure the effectiveness of sampling equivariance of SES-Net using these works. Thus, we used Gliding~\cite{xu2020gliding} as our baseline model, which has a relatively simple design and excellent performance. We also conducted experiments on the state-of-the-art model S$^2$ANet combined with our SES-Layer to verify the versatility of our method.

We compared the performance of our proposed Gliding+SES and S$^2$ANet+SES with FR-O~\cite{xia2018dota}, Azimi et al.~\cite{azimi2018towards}, CADNet~\cite{zhang2019cad}, DRN~\cite{pan2020dynamic}, R$^3$Det~\cite{yang2019r3det}, SCRDet~\cite{yang2019scrdet}, CenterMap~\cite{wang2020learning}, Li et al.~\cite{li2019learning}, Gliding~\cite{xu2020gliding} and S$^2$ANet~\cite{han2021align}; results are shown in Tab. \ref{dota}. 
In Tab. \ref{para}, we also compare the number of backbone parameters and backbone FLOPs  with the baseline models. 
Compared to Gliding, Gliding+SES has significantly better mAP, using fewer parameters and FLOPs: the better sampling equivariance provided by our proposed model has clear advantages for tasks involving a wide range of transformations like aerial image object detection. 

We also conducted experiments using the state-of-the-art model S$^2$ANet with our proposed SES-Layer. 
S$^2$ANet uses an alignment convolution layer (ACL) in the detection head to handle rotation and scaling of the object, which partially overlaps and conflicts with the function of the proposed SES-Layer. Even so, our experiments  show that S$^2$ANet+SES improves mAP over existing SOTA models using fewer parameters and FLOPs, indicating the robustness and effectiveness of our proposed method.
Various detection results using Gliding and our proposed Gliding+SES are shown in Fig. \ref{dotaexp}: our method both has a higher detection rate, and provides more accurate bounding boxes.

\begin{table}[!t]
	\begin{center}
		\caption{Ablation experiments using the  DOTA test set.  }
		\label{ablation}
		\begin{tabular}{llllllll}
			\hline\noalign{\smallskip}
			    Method & mAP & B-Params & B-FLOPs \\
			\noalign{\smallskip}
			\hline
			\noalign{\smallskip}
        		Gliding & 74.82 & 44.5M & 7.8G\\
        		Gliding+DCNv2 & 75.26 & 44.9M & 7.8G\\
        		Gliding+SA & 75.81 & 38.4M & 7.6G\\
        		Gliding+SA+RNM & 75.84 & 38.4M & 7.6G\\
        		\noalign{\smallskip}
        		\hline
        		\noalign{\smallskip}
        		Gliding+SES w/o RBN & 76.10 & 38.5M & 7.6G\\
        		\noalign{\smallskip}
        		\hline
        		\noalign{\smallskip}
        		Gliding+SES(Ours) & \textbf{76.84} & 38.5M & 7.6G\\
			\hline
		\end{tabular}
	\end{center}
\end{table}

\begin{table}[!t]
	\begin{center}
		\caption{Ablation experiments using the  ImageNet test set.  }
		\label{ablation2}
		\begin{tabular}{llllllll}
			\hline\noalign{\smallskip}
			    Method & Top-1(\%) & Top-5(\%) \\
			\noalign{\smallskip}
			\hline
			\noalign{\smallskip}
			    SES-Net10+RNM & 76.24 & 92.98\\
			    SES-Net10 & 76.40 & 93.04\\
			\hline
		\end{tabular}
	\end{center}
\end{table}

\subsection{Ablation Studies}

In order to verify the effectiveness of our proposed modules and explore other possible network configurations, we conducted a series of ablation studies.

\subsubsection{Validation of SES-Layer}
We replaced the sampling equivariance module with other modules in Gliding to verify its effectiveness. We replaced [RNM, ReLU, SES-Layer] in Gliding+SES with [RNM, ReLU, SA-Layer] (denoted Gliding+SA+RNM), [BN, ReLU, SA-Layer] (denoted Gliding+SA), [BN, ReLU, Conv3x3] (denoted  Gliding), and [BN, ReLU, DCNv2] (denoted Gliding+DCNv2), where SA-Layer is the self-attention layer of SAN. These models were all initialized with pretrained weights on ImageNet.  We tested all models on DOTA, with results  shown in  Tab. \ref{ablation}.
Gliding+SES achieves the best performance, while the number of parameters and FLOPs remains similar or even smaller,  showing that the sampling equivariance of the SES-Layer can extract more powerful features from objects undergoing various transformations, using the same number of network parameters.

\subsubsection{Validation of Randomized Normalization Module}
To verify the effectiveness of our proposed RNM module, we conducted ablation studies by replacing the RNM of Gliding+SES with BN (denoted  Gliding+SES w/o RBN) using DOTA. Results are shown in Tab. \ref {ablation}, and show that our proposed RNM module enables the model to generate more generalized sampling masks and enhance the generalization ability of the model, providing significantly improved results on a relatively small-scale dataset.
We also considered replacing [BN, ReLU, SES-Layer] in SES-Net10 with [RNM, ReLU, SES-Layer] (denoted  SES-Net10+RNM), and tested the classification accuracy on ImageNet. As  Tab. \ref{ablation2} shows, due to the relatively large amount data in ImageNet, RNM has no significant impact on the results.

We further conducted experiments with different configurations of RNM.
We tried various combinations of input features with or without random disturbance in  $Q$, $K$ and $V$ in the SES-Layer. For example, we tried random disturbance $X''$ for $Q$ and  without random disturbance $X'$ for $K$ and $V$. This model is denoted  Gliding+SES-Random-Q. Likewise, other combinations are respectively denoted  Gliding+SES-Random-K, Gliding+SES-Random-V, Gliding+SES-Random-QK, Gliding+SES-Random-QV, Gliding+SES-Random-KV and Gliding+SES-Random-QKV.
Results are shown in  Tab. \ref{more}. We can see that Gliding+SES-Random-QK provides the best performance on DOTA,  verifying our hypothesis that the query and key of the self-attention structure can easily overfit the training data, making it hard to generate more generalized sampling masks during inferencing. Therefore, random disturbance should be added to the query and key input features. Adding random disturbance to the value will increase  noise in those features, which is not conducive for training.

We also conducted a series of experiments to explore the effects of different random size $r=0.001, 0.005, 0.02, 0.08, 0.32$ for Gliding+SES. Results are shown in Tab. \ref{more}, leading us to conclude the optimal  value in  our experiments is $r=0.005$.

\begin{table}[!t]
	\begin{center}
		\caption{Further experiments using the  DOTA test set.  }
		\label{more}
		\begin{tabular}{llllllll}
			\hline\noalign{\smallskip}
			    Method  & mAP \\
			\noalign{\smallskip}
			\hline
			\noalign{\smallskip}
        		Gliding+SES-$r=0.001$ & 75.86\\
        		Gliding+SES-$r=0.005$ & \textbf{76.84}\\
        		Gliding+SES-$r=0.02$ & 76.45\\
        		Gliding+SES-$r=0.08$ & 75.72\\
        		Gliding+SES-$r=0.32$ & 75.80\\
        		\noalign{\smallskip}
        		\hline
        		\noalign{\smallskip}
        		Gliding+SES-Random-QKV & 75.23\\
        		Gliding+SES-Random-QK & \textbf{76.84}\\
        		Gliding+SES-Random-QV & 74.02\\
        		Gliding+SES-Random-KV & 75.21\\
        		Gliding+SES-Random-Q & 75.78\\
        		Gliding+SES-Random-K & 75.87\\
        		Gliding+SES-Random-V & 75.74\\
			\hline
		\end{tabular}
	\end{center}
\end{table}
\section{Conclusions}
Sampling equivariance is an important property, which can enhance the expressive ability of a model without increasing model parameters.
In this paper, we analyze why existing methods  do not have complete sampling equivariance, and propose a new solution, SES-Net. 
Through quantitative comparisons, we demonstrate that our proposed method has significantly better sampling equivariance than existing methods, and performs better  than alternatives without increasing the number of model parameters or computational effort. We also propose an RNM module to enhance the generalization ability of the model by adding random disturbances to part of the data without changing the data distribution.

Object detection in aerial images has greater requirements for sampling equivariance due to the characteristics of overhead shooting. We have applied our proposed SES-Layer to existing aerial image object detection methods \cite{xu2020gliding, han2021align} to provide better sampling equivariance. 
Experiments show that RNM can enhance the generalization ability of our proposed network, and that it works well with the SES-Layer to achieve state-of-the-art performance on the DOTA benchmark without computational overhead. 

In principle, our approach can be used in many popular network architectures, benefiting various computer vision tasks, which we will further consider in future.

\bibliographystyle{IEEEtran}
\bibliography{egbib}


%

\ifCLASSOPTIONcompsoc
  \section*{Acknowledgments}
\else
  \section*{Acknowledgment}
\fi

This work was supported by the Natural Science Foundation of China (Project Number 61521002), Research Grant of Beijing Higher Institution Engineering Research Center and Tsinghua-Tencent Joint Laboratory for Internet Innovation Technology. We would like to thank Yuan-Chen Guo for helpful discussions and help in writing, and Dr. Dun Liang for helpful discussions.

\ifCLASSOPTIONcaptionsoff
  \newpage
\fi

\begin{IEEEbiography}[{\includegraphics[width=1in,height=1.25in,clip,keepaspectratio]{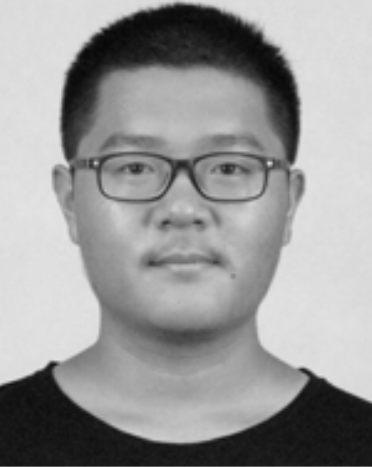}}]{Guo-Ye Yang}
received a B.S. degree from Tsinghua University in 2019, where he is currently pursuing a Ph.D. degree. His research interests include computer graphics, image analysis, and computer vision.
\end{IEEEbiography}

\begin{IEEEbiography}[{\includegraphics[width=1in,height=1.25in,clip,keepaspectratio]{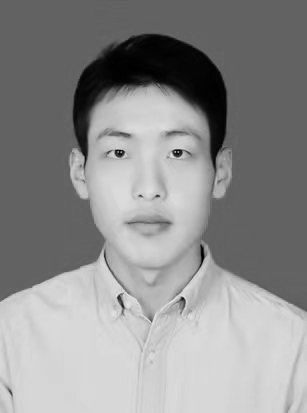}}]{Xiang-Li Li}
received a B.S. degree from Dalian University of Technology in 2019; he is currently pursuing a Ph.D. degree in Tsinghua University. His research interests include computer graphics, image analysis, and computer vision.
\end{IEEEbiography}


\begin{IEEEbiography}[{\includegraphics[width=1in,height=1.25in,clip,keepaspectratio]{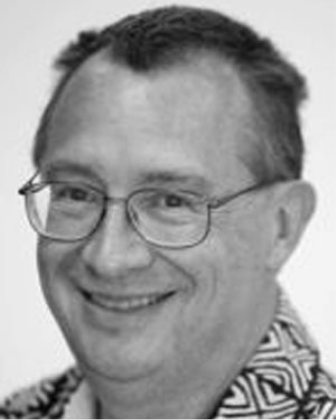}}]{Ralph R. Martin}
received his Ph.D. degree from Cambridge University in 1983. He is currently a emeritus professor with Cardiff University. He has authored over 250 papers and 14 books, covering such topics as solid and surface modeling, intelligent sketch input, geometric reasoning, reverse engineering, and various aspects of computer graphics. He is a Fellow of the Learned Society of Wales, the Institute of Mathematics and its Applications, and the British Computer Society. He is currently the Associate Editor-in-Chief of Computational Visual Media.
\end{IEEEbiography}

\begin{IEEEbiography}[{\includegraphics[width=1in,height=1.25in,clip,keepaspectratio]{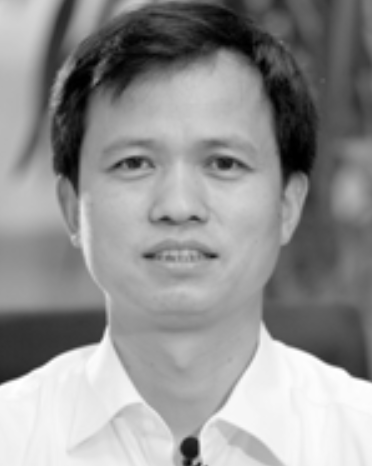}}]{Shi-Min Hu}
received  a  Ph.D.  degree  from Zhejiang University in 1996. He is currently a Professor with the Department of Computer Science and Technology, Tsinghua University, Beijing. He has published over 100 articles in journals and refereed conferences. His research interests include digital geometry processing, video processing, rendering, computer animation, and computer-aided geometric design. He is the Editor-in-Chief of Computational Visual Media, and is on the editorial board of several journals, including  Computer Aided Design, and Computers and Graphics.
\end{IEEEbiography}




\end{document}